\pdfoutput=1

\documentclass[runningheads]{llncs}
\usepackage[T1]{fontenc}
\usepackage{graphicx}
\usepackage{xcolor}

\usepackage{todonotes}

\usepackage{amsmath}

\newcommand{\rhomax}{\rho^{max}}
\newcommand{\rhomaxmax}{\rho^{N-1}}
\newcommand{\dnthreshold}{\eta}
\newcommand{\riskthreshold}{\theta}

\usepackage{tabularx}
\usepackage{threeparttable}
\usepackage{multirow}

\usepackage{subfig}

 \usepackage{algorithm}
\usepackage[noend]{algcompatible}
\algrenewcommand\algorithmicindent{2.0em}
\makeatletter
\newlength{\trianglerightwidth}
\algnewcommand{\LineComment}[1]{\State $\triangleright$ #1}
\algnewcommand{\LineCommentCont}[1]{\State \hskip\ALG@thistlm%
  \parbox[t]{\dimexpr\linewidth-\ALG@thistlm}{\hangindent=\trianglerightwidth \hangafter=1 \strut$\triangleright$ #1\strut}}

\usepackage[hyphens]{url}
\usepackage{hyperref}

\usepackage{upgreek}

\begin{document}
\title{Imitation Learning for Intra-Day Power Grid Operation through Topology Actions}
\titlerunning{Imitation Learning for Intra-Day Power Grid Operation}
%
%
\author{Matthijs de Jong \inst{1,2}$^{,*}$\orcidID{0009-0006-9641-5092} \and
Jan Viebahn\inst{2}\orcidID{0000-0003-3118-7691} \and
Yuliya Shapovalova\inst{1}\orcidID{0000-0002-4487-5947}}
\authorrunning{M. de Jong et al.}
%
\institute{
Radboud University, 6525 XZ  Nijmegen, The Netherlands \\ \email{\{matthijs.dejong,yuliya.shapovalova\}@ru.nl} \and
TenneT TSO, 6812 AR Arnhem, The Netherlands  \\\email{Jan.Viebahn@tennet.eu} \\
$^*$ corresponding author}
\maketitle              

\begin{abstract}
Power grid operation is becoming increasingly complex due to the increase in generation of renewable energy. The recent series of Learning To Run a Power Network (L2RPN) competitions have encouraged the use of artificial agents to assist human dispatchers in operating power grids. 
In this paper we study the performance of imitation learning for day-ahead power grid operation through topology actions. In particular, we consider two rule-based expert agents: a greedy agent and a N-1 agent. While the latter is more computationally expensive since it takes N-1 safety considerations into account, it exhibits a much higher operational performance. 
We train a fully-connected neural network (FCNN) on expert state-action pairs and evaluate it in two ways. First, we find that classification accuracy is limited despite extensive hyperparameter tuning, due to class imbalance and class overlap. Second, as a power system agent, the FCNN performs only slightly worse than expert agents. Furthermore, hybrid agents, which incorporate minimal additional simulations, match expert agents' performance with significantly lower computational cost. Consequently, imitation learning shows promise for developing fast, high-performing power grid agents, motivating its further exploration in future L2RPN studies.

\end{abstract}

\section{Introduction}

The energy transition is a crucial development for ensuring the sustainability and future security of society.
Transmission system operators (TSOs) play a crucial role in this transition and are consequently faced with new challenges.
Consumption pattern changes, such as those caused by electrical vehicles, are rapidly increasing the energy demand.
In the Netherlands, the energy consumption is expected to grow thirty percent between 2024 and 2033; the projections are similar for other European countries \cite{tennet_2024}.
Simultaneously, renewable energy comes from sources that are often uncontrollable and more variable.
The percentage of energy from controllable sources is expected to fall by approximately forty by 2033 in the Netherlands \cite{tennet_2024}. 

Congestion management is one of these challenges \cite{f7930ddbf547488bafa56488ef974ad0}.
Exploiting the flexibility in the network topology presents an opportunity for remedying grid congestion \cite{jan_viebahn_2024}.
However, the topology space is combinatorially large, and hence too difficult to explore by humans and too computationally expensive for traditional computational methods \cite{f7930ddbf547488bafa56488ef974ad0}.

Recent advancements in machine learning (ML) present new opportunities for grid topology control \cite{f7930ddbf547488bafa56488ef974ad0}. By being exposed to a vast number of grid states and topologies, machine learning models can learn successful combinations between them. 
Moreover, machine learning provide greater computational efficiency compared to traditional algorithmic paradigms, which is essential for solving complex problems like topology control. 
Operating the power grid is a sequential decision making problem, for which the two main ML paradigms are imitation learning (IL) and reinforcement learning (RL).
With IL, a model learns from the state-action pairs from an expert agent, whereas with RL, the agent learns from directly interacting with the environment \cite{5480345}. 
So far, most research has focused on RL methods for grid topology control \cite{yoon2020winning,pan2021improving,zhou2021action,Subramanian_2021}.
RL methods have the benefit that they can learn behavior beyond what the expert agents exhibit.
IL methods have only been applied shallowly for pre-training models for RL, which helps RL agents learn complex behavior \cite{lan2019aibased,githubGitHubAsprinChinaL2RPN_NIPS_2020_a_PPO_Solution,lehna2023managing}.
No research has investigated IL as the main method for developing topology control agents, despite the fact that it has certain benefits over RL.
Training with IL is more simple and efficient, and because the model learns to imitate expert agents, there is less uncertainty about the model's actions.

In this work, we investigate rule-based expert agents and ML agents that are based on IL.
We consider two rule-based agents: a greedy agent and a N-1 agent that aims to satisfy N-1 redundancy.
We specifically look at grid control in an intra-day setting, instead of the more commonly used but less realistic intra-month period.
We also consider various regimes of line outages.
We perform an extensive hyperparameter tuning and error analysis to the ML models, and apply them to the grid directly or as hybrid agents.
This research aims to investigate the merit that imitation learning and hybrid agents have at operating the grid.

The remainder of the paper is organized as follows. In section 2 we discuss related work. Section 3 describes the power grid setup and experimental setting. Section 4 describes the rule based agents as well as the results relating to them. Section 5 describes the IL agents and the related results. Section 6 provides discussion and a summary, and section 7 directions for future research. Additionally, Table \ref{tab:notation} lists the symbols and notation used throughout the paper.
The code for this project is shared on Github \footnote{\url{https://github.com/MatthijsdeJ/Imitation_Learning_Topology_Control}}.

\begin{table*}[t]
\centering
\caption{
Mathematical symbols and notation used in this paper, and their meaning.
}
\label{tab:notation}
\begin{tabularx}{\textwidth}{|l|X|} 
\hline
\textbf{Notation} & \textbf{Meaning} \\
\hline
$O$ & The set of grid objects (i.e. generators, loads, and line endpoints).  \\ \hline
$S$ & The set of substations. \\ \hline
$L$ & The set of power lines. \\ \hline
$A$ & The space of valid, unique set-actions.  \\ \hline
$\rho_{l}$ & The loading, i.e. the current over the thermal limit, of line $l \in L$.\\ \hline
$\rhomax = \mathrm{max}_{l \in L} \rho_{l}$ & The \textit{maximum loading}, i.e. the loading of the highest-loaded line. \\ \hline
$(\rhomax|a)$ & The maximum loading resulting from simulating action $a \in A$. \\ \hline
$(\rhomax|a,l)$ & The maximum loading resulting from simulating action $a \in A$ and the disablement of line $l \in L$.  \\ \hline
$\rhomaxmax$ & The \textit{maximum N-1 loading} of the power network, see \S \ref{sssec:N-1_agent}.\\ \hline
$\dnthreshold$ = 0.97 & The activity threshold parameter. \\ \hline
$\riskthreshold = 1.0$ & The risk threshold parameter, used by various agents. \\ \hline
$\alpha = 0.1$ & The label weight hyperparameter. \\ \hline
$\textbf{p} = (0,1)^{|O|}$ & The output of a ML model.  \\ \hline
$\textbf{y} = \{0,1\}^{|O|}$ & The target of a ML model. \\ \hline
$\textbf{w} = \{\alpha,1\}^{|O|}$ & The label weights. \\ \hline
$ \{\textbf{p}_s\} {} \forall s \in S$ & A partition of the output vector $\textbf{p}$ into subvectors $\textbf{p}_s$ corresponding to the object predictions at substation $s \in S$.  \\ \hline
\end{tabularx}
\end{table*}

\section{Related Work}
\label{sec:related_work}

The L2RPN competition was premiered in 2019 to foster research between the power system and ML communities.
The first competition focused on operating scenarios of the IEEE14 network within a specific time frame \cite{marot2019learning}.
The second L2RPN competition was featured at NeurIPS 2020 \cite{marot2021learning}.
It featured the much larger IEEE118 bus network, and two tracks: a robustness track with unplanned outages, and an adaptability track with evaluation on unseen generation/load distributions.
The submissions were subsequently analyzed \cite{marot2021learning}. 
The biggest takeaway is the importance of combining ML and simulation functionality, which was done by four out of five of the top submissions on both tracks.
Interestingly, the top five submissions of both tracks also contained one purely rule-based agent which simulated pre-computed subsets of topological actions \cite{marot2021learning}.
The authors also emphasize the importance of discovering robust actions and the ability to plan sequences.
Following competitions focused on trust and the sending of an alarm signal \cite{marot2022learning}, and the inclusion of batteries \cite{serré2022reinforcement}.

Lan et al. \cite{lan2019aibased} presented the first imitation learning approach in 2019.
They pre-train a deep Q-network with an expert power grid simulator.
This approach was expanded upon by Binbinchen in their 2020 L2RPN submission \cite{githubGitHubAsprinChinaL2RPN_NIPS_2020_a_PPO_Solution}.
They introduce a curriculum framework, consisting of four modules which sequentially perform: action space reduction, rule-based greedy operation, imitation learning, and reinforcement learning.
Lehna et al. \cite{lehna2023managing} refine this framework by introducing topology reversal and a higher-performing N-1 expert agent.
In recent research, Lehna et al. \cite{lehna2024hugo} develop this direction further by using an agent that combines the curriculum framework with a heuristic target topology approach.
The winning submission of the 2023 L2RPN competition also featured a model developed with this curriculum framework \cite{javaness_2023}.

\section{Power Grid Setup}
\label{sec:power_grid_setup}

\begin{figure}[t]
\centering
\includegraphics[width=0.85\textwidth]{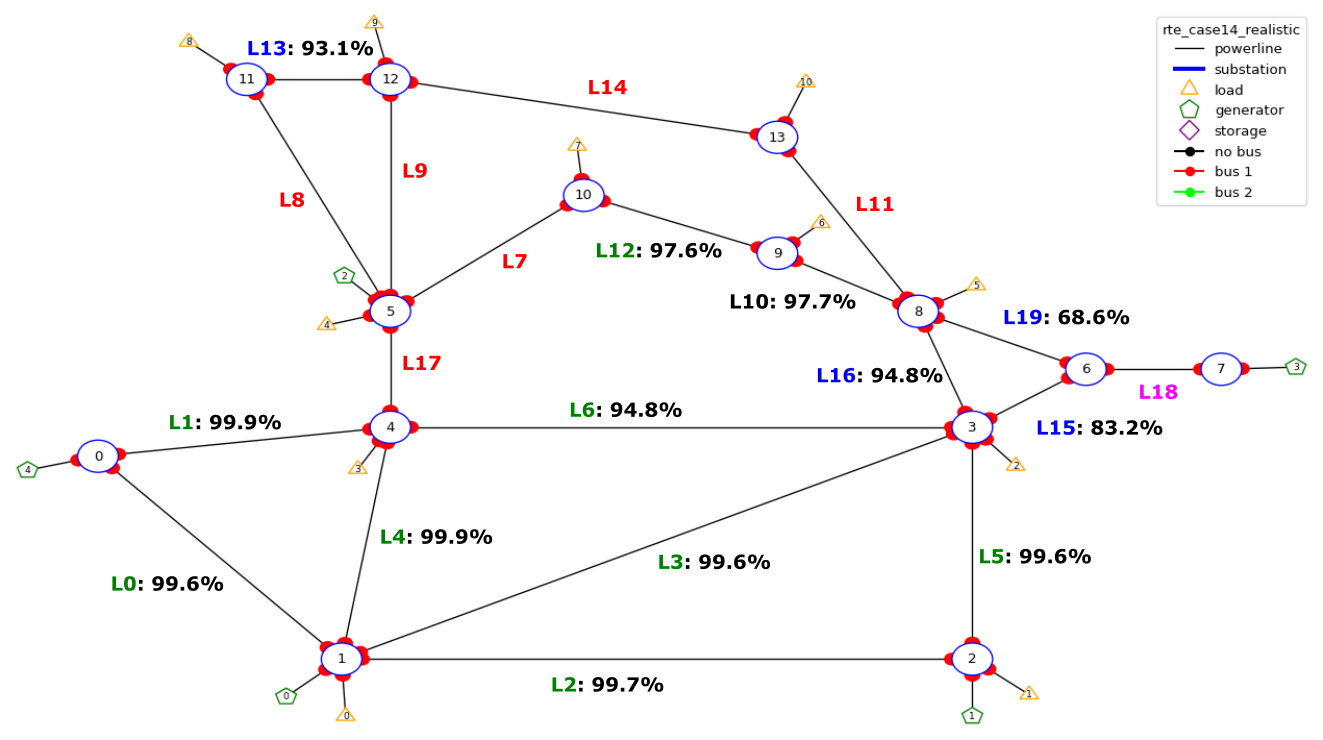}
\caption{The default state of Grid2Op environment \texttt{rte\_case14\_realistic}.
The percentages refer to the percentage of days completed of the greedy agent on the N-1 network with the annotated line disabled (read \S\ref{sssec:N-1_networks}).
Lines annotated with green denote the first cluster and lines annotated with blue the second.
The Greedy agent could not operate well with the lines annotated with red disabled. \protect\footnotemark
Line 10 (black) does not clearly fall in any cluster, and line 18 (pink) cannot be disabled, as the resulting topology is inherently invalid.}
\label{fig:rte_case14_realistic_annotated}
\end{figure}

\sloppy
We employ the IEEE 14-bus system using the \texttt{rte\_case14\_realistic}\footnote{This environment has since been deprecated, future research should use environment \texttt{l2rpn\_case14\_sandbox}.} environment of the Python package Grid2Op\footnote{\url{https://grid2op.readthedocs.io/}}.
The power grid is shown in Figure \ref{fig:rte_case14_realistic_annotated} and includes fourteen substations, five generators, eleven loads, and twenty power lines. 
The generators consist of one solar, one nuclear, one wind-based, and two thermal generators.
The grid is divided into two sides: the high-voltage transmission side containing substations 0 to 4, and the low-voltage distribution side containing substations 5 to 13. 
Lines 15 to 19 model the transformers connecting these sides.
We adjust the thermal limits of the lines to make differences between the transmission and distribution parts more pronounced and realistic, as described by Subramanian et al. \cite{Subramanian_2021}.

Electric current flows through power lines.
Each line has a thermal limit, which the load cannot exceed for an extended period of time, lest the line fails and is disabled.
The ratio of the current to the thermal limit of a line is called its \textit{loading}, denoted by $\rho$ (see Table \ref{tab:notation}). Electricity previously flowing through a failed line finds a new path through the network, increasing the load of the power lines on that path.
That increases the risk of those lines failing, potentially causing a cascade of failing lines. 
Cascading failures can render parts of or the entire power grid nonfunctional.
The aim of topology control is to avoid such failure by redirecting current flow. 
Power networks are also designed to avoid cascading failures by satisfying \textit{N-1 redundancy}: the property that the network remains stable in the presence of the failure of any single line.
We investigate the ability of agents to prepare for, and control during, line outages.
To that end, we also consider variations of the network with a single line disabled, which we call \textit{N-1 networks}.

\subsection{Action space}
In this study, we focus on topological remedial actions.
Topology control is possible because each substation contains two busbars to which grid objects (these are generators, lines, and line endpoints) can be attached.
Electricity flows between objects attached to the same busbar and between the two endpoints of a power line, but not between objects on different busbars within the same substation.
As such, reconfiguring the object-busbar attachments allows control over the flow of electricity in the network.
In Grid2Op, the \textit{topology vector} variable indicates at which busbar of its substation the object is attached.
Values of 1, 2, and -1 represent attachment to the first and second busbar, and disablement, respectively.
\footnotetext{These percentages are not displayed, as runs that could complete fewer than 5\% of the first 140 days were terminated early.}

In Grid2Op, topology control actions correspond to changing object-busbar attachments through either absolute set-actions or relative switch-actions.
Grid2Op introduces real-world constraints by allowing object-busbar changes at only a single substation per timestep.
Certain configurations of object-busbar attachments at substations are either invalid or redundant. 
For instance, actions that isolate generators or loads are considered invalid in Grid2op.
Similarly, configurations mirrored w.r.t. busbar attachments are redundant. 
We use the approach by Subramanian et al. \cite{Subramanian_2021} to calculate the space of valid, unique set-actions (i.e., substation configurations).
Note that as set-actions are represented absolutely, set-actions might set the busbar configurations to states they are currently in, effectively being do-nothing actions.
We call these \textit{implicit} do-nothing actions, in contrast to the \textit{explicit} do-nothing actions described in \S \ref{sssec:greedy}.

\subsection{Intra-day Scope \& Regimes}
\label{ssec:scope_regime}
The environment features one thousand scenarios, each consisting of 8064 timesteps separated by intervals of five minutes, i.e. 28 days.
Each scenario features different generation and load profiles. 
Because of the intra-day setup, we split the scenarios into individual days.
Between days, the topology is reset to the default topology, in which all enabled objects are attached to the first busbar.
Besides realism, this has two benefits: (1) shortening the duration of runs increases the amount of usable data for imitation learning, as data from runs where the expert agent fails cannot be used, and (2) this emulates topology reversal, which has been found to be beneficial for operating power grids \cite{lehna2023managing}.
A game-over occurs when the power grid fails to transport sufficient power from generators to loads due to line failures.
To avoid circular analysis, we split the scenarios into 70/10/20\% train/validation/test sets.
We excluded three scenarios where Grid2Op failed to converge the power flow without agent misoperation.
These scenarios have been omitted throughout the research.

We consider three different regimes which represent different difficulty levels for grid operation.
The \textit{full-network} regime considers the full network, i.e. without outages. 
In the \textit{planned-outage} regime, a single line is disabled for the entire day.
The \textit{unplanned-outage} regime introduces an opponent that disables lines randomly. 
In specifying the opponent we follow Blazej et al. \cite{manczak2023hierarchical}. The opponent disables a uniformly sampled line for four hours twice a day. 
We use lines 0, 2, 4, 5, 6, and 12 as the lines that the opponent can disable to match the disabled lines in the dataset (read \S \ref{ssec:dataset}).
We include a cooldown period, so that the outages are separated by minimally an hour.

\section{Rule-based Expert Agents}
\label{sec:rule_based_agents}

\subsection{Greedy Agent}
\label{sssec:greedy}
The greedy agent simulates each action and selects the action that minimizes the maximum loading $\rhomax$.
To avoid hyperactive behavior, we introduce an activity threshold $\dnthreshold$ \cite{Subramanian_2021}, which we set to 0.97.
The agent selects an explicit do-nothing action if the maximum loading does not exceed this threshold. 
Algorithm \ref{alg:greedy} presents the pseudocode for the greedy agent.

\begin{algorithm}
\caption{Algorithm for the greedy agent.}
\label{alg:greedy}
\textbf{Input:} Maximum loading $\rhomax$, simulated maximum loadings $(\rhomax|a)$ \\
\textbf{Output:} Selected action $a \in A$
\begin{algorithmic}[1]
\LineComment{If no current risks, return do-nothing.}
\IF {$\rhomax < \dnthreshold$} 
    \State \textbf{return} 
$a_{do\_nothing}$ 
\ENDIF
\LineComment{Select the action that minimizes $\rhomax$.}
\State \textbf{return} $\mathrm{argmin}_{a \in A} \ (\rhomax|a)$
\end{algorithmic}
\end{algorithm}

\subsection{N-1 Agent}
\label{sssec:N-1_agent}
The second rule-based agent aims to achieve grid configurations that satisfy N-1 redundancy, i.e. grid configurations robust to the disablement of single lines.
This agent quantifies the N-1 risk of the resulting topologies and selects the action that minimizes this risk.

N-1 risk is quantified by the \textit{maximum N-1 loading}, denoted $\rhomaxmax$.
This is calculated as the maximum of the maximum loadings, $\rhomax$, among the N-1 networks.
Evaluation of the greedy agent on the N-1 networks showed that the network became almost entirely inoperable when certain lines were disabled (see \S\ref{sssec:N-1_networks}.
Therefore, we only consider the disablement of a subset of lines, $L^{disable} \subset L$.
The excluded lines are annotated with red or pink in Figure \ref{fig:rte_case14_realistic_annotated}.
Thus, the maximum N-1 loading for the topology resulting from simulating a particular action $a$ is given by
\begin{align}
\rhomaxmax(a) = \mathrm{max}_{l \in L^{disable}} (\rhomax|a,l) 
\nonumber \end{align}
where $(\rhomax|a,l)$ is the maximum loading resulting from simulating action $a$ and the disablement of line $l$.
In case of a game-over, we consider $\rhomax = \infty$. 
Therefore, if disabling one of the lines leads to an immediate game-over, $\rhomaxmax = \infty$.
Algorithm \ref{alg:n-1_pseudocode} presents the pseudocode for the N-1 agent.
The policy of the N-1 agent consists of three steps:

\begin{enumerate}
    \item \textbf{Activity check}: Similarly to the greedy agent, if the maximum loading  does not exceed the inactivity threshold ($\rhomax < \dnthreshold$), an explicit do-nothing action is selected.
    \item \textbf{Minimizing N-1 risk}: Above that threshold, the agent simulates each action and considers the subset of actions $A_{secure} \subset A$ that do not put the simulated network at risk:
    \begin{align}
        A_{secure} = \{a \in A | (\rhomax|a) <  \riskthreshold\} ,\nonumber
    \end{align}
    where $\riskthreshold = 1.0$ is the risk threshold parameter.
    Among $A_{secure}$, the action with the lowest maximum N-1 loading $\rhomaxmax$ is found:
    \begin{align}
        a_{best} = \mathrm{argmin}_{a \in A_{secure}} \ \rhomaxmax(a). \nonumber
    \end{align}
    If the lowest finite maximum N-1 loading is finite, i.e. $\rhomaxmax(a_{best}) < \infty$, then the corresponding action is selected.
    \item \textbf{Fallback to greedy behavior}: If either $A_{secure}$ is empty or the lowest maximum N-1 loading is non-finite, then an action is taken greedily instead.
\end{enumerate} 
Our algorithm differs from the one used by Lehna et al. \cite{lehna2023managing} through the addition of the pre-selection for secure actions and the greedy-fallback mechanism. 
In the unplanned-outage regime, the N-1 agent is replaced by the Greedy agent during unplanned outages.

\begin{algorithm} 
\caption{Algorithm for the N-1 agent.} 
\label{alg:n-1_pseudocode} 
\textbf{Input:} Maximum loading $\rhomax$, simulated maximum loadings $(\rhomax|a)$ and $(\rhomax|a,l)$ \\
\textbf{Output:} Selected action $a \in A$
\begin{algorithmic}[1] 
    \LineComment Define the N-1 max loading function.
    \STATE \textbf{def} $\rhomaxmax(a)$: $\mathrm{max}_{l \in L^{disable}} (\rhomax|a,l) $
    
    \LineComment{If no current risks, return do-nothing.}
    \IF {$\rhomax < \dnthreshold$}
        \STATE \textbf{return} $a_{do\_nothing}$
    \ENDIF

    \LineComment{Try to find a suitable N-1 redundant topology.}
    \State $A_{secure} \gets \{a \in A | (\rhomax|a) < \riskthreshold\}$
    \IF{ $A_{secure} \neq \emptyset$}

        \State $a_{best} \gets \mathrm{argmin}_{a \in A_{secure}} \rhomaxmax(a)$
        \State $\rho_{best}^{N-1} \gets \rhomaxmax(a_{best})$

        \LineComment{Return the best action if it leads to a N-1 redundant grid.}
        \IF{$\rho^{N-1}_{best} \neq \infty$} 
                \State \textbf{return} $a_{best}$
        \ENDIF
    \ENDIF

    \LineComment{Select an action greedily.}
    \State \textbf{return} $\mathrm{argmin}_{a \in A} \ (\rhomax|a)$
\end{algorithmic}
\end{algorithm}

\subsection{Results of the Rule-based Agents}

\subsubsection{Full-Network Regime.}
\label{ssec:full_network_regime}
For the generation of data, both rule-based agents were applied to all scenarios of the full-network regime. 
Table \ref{tab:agent_performance}, column 2, summarizes the performance of the greedy and N-1 agent.
The greedy agent completed nearly all days (99.7\%) and the N-1 agent all days.

\makeatletter
\newcommand{\thickhline}{%
    \noalign {\ifnum 0=`}\fi \hrule height 1pt
    \futurelet \reserved@a \@xhline
}
\begin{table}[tb]
\begin{threeparttable}[t]
\caption{
The percentage of days completed of the agents in various regimes (columns 2-5), and the mean inference speed (column 6).
The agents were evaluated on the test scenarios.
The results in the fifth column are averaged over five seeds of random outages.
In fourth to eighth row, the results are averaged over the five FCNN models.
In their intersection, there results are averaged over five different runs, each with a different model and random outages.
The sixth represents the mean of the inference duration of the various agents on the first fifty validation chronics.}
\begin{tabular}{|p{0.155\linewidth}|p{0.155\linewidth}|p{0.155\linewidth}|p{0.155\linewidth}|p{0.155\linewidth}|p{0.155\linewidth}|}
\hline
& \multirow{2}{=}{Full network (\%)}  & 
\multirow{2}{=}{Planned Line 2 Outage(\%)}  &
\multirow{2}{=}{Planned Line 6 Outage(\%)}  &
\multirow{2}{=}{Unplanned Outage (\%)}  &
\multirow{2}{=}{Duration ($\upmu$s)}  \\ & & & & & \\ \thickhline
Do-nothing & 59.80 & 54.95 & 23.70                & 46.57±0.17 & -\tnote{a} \\ \hline
Greedy & 99.73\tnote{b}& 99.70\tnote{b} & 94.77\tnote{b} & 81.79±1.02 & 6.77E5 \\ \hline
N-1 & \textbf{100.00}\tnote{b} & -\tnote{c}  & -\tnote{c} & 92.39±0.24\tnote{d} & 4.70E6 \\
\thickhline
Naive & 96.27±0.90 & 97.69±0.25 & 88.18±0.99 & 86.33±0.90 & 5.74E2 \\ \hline
Verify & 98.95±0.19 & 98.94±0.32 & 90.36±1.07 & 89.62±0.57 & 1.14E4 \\ \hline
Ver.+Greedy & 99.85±0.04 & \textbf{99.82}±0.03 & \textbf{97.00}±0.41 & 93.76±0.16 & 3.29E4 \\ \hline
Verify+N-1 & 99.97±0.02 & -\tnote{c} & -\tnote{c} & \textbf{95.17}±0.18\tnote{d} & 9.26E4 \\ \hline
\end{tabular}
\tnote{a}Omitted because the duration of the do-nothing agent should be near-instantaneous. 
\tnote{b}These results were computed over all scenarios as part of data analysis.
\tnote{c}Omitted because the N-1 agent cannot be applied to N-1 networks.
\tnote{d}During unplanned line outages, the N-1 agent was replaced by the greedy agent.
\label{tab:agent_performance}
\end{threeparttable}
\end{table}

Notably, the N-1 agent takes more actions than the greedy agent (51,716 and 25,843 actions in total, respectively).
This can be explained by the greedy agent pursuing a lower maximum loading more aggressively than the N-1 agent:
the N-1 agent more often selects actions that do not lower the maximum loading below the do-nothing threshold, keeping it in an active state.
Respectively 7.5\% and 36.4\% of the actions of the greedy and N-1 agent were followed by another action, indicating that the N-1 agent has longer action sequences.

\begin{figure}[t]%
    \centering
    \subfloat{\includegraphics[width=6cm]{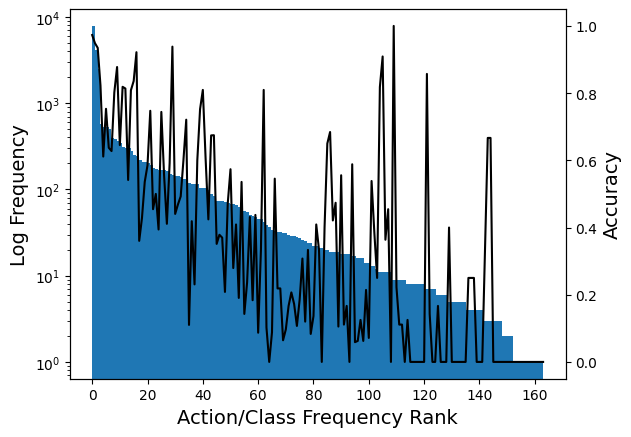} \label{fig:accuracy_per_label}}%
    \subfloat{\includegraphics[width=5.5cm]{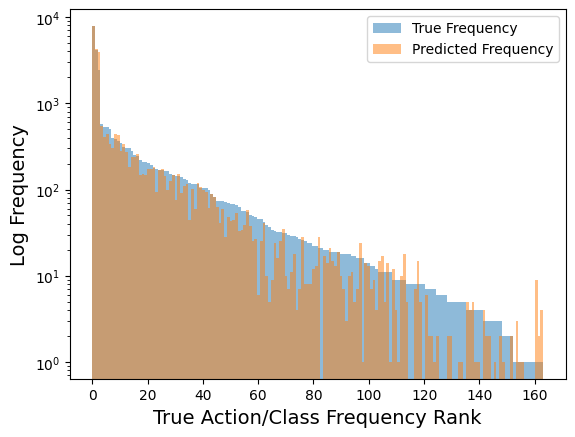}
    \label{fig:label_predictions}}
    \caption{
    \textbf{Left}: A log Pareto chart of the actions/classes in the validation set (blue) and corresponding accuracy per action/class (black). 
    A negative trend between frequency rank and accuracy can be observed.
    \textbf{Right}: A log Pareto chart of the actions/classes in the validation set (blue) and how often they were predicted by the ML model (orange).
    The blue area that does not overlap with the orange area at higher ranks indicates that the model is biased against rare classes.
    }
    \label{fig:example}%
\end{figure}

Another noteworthy aspect is the imbalanced distributions of both actions and resulting topologies. 
A few actions and topologies dominate the distribution; particularly so for the Greedy agent.
The imbalanced action distribution can be observed in the combined dataset in Figure \ref{fig:accuracy_per_label}.

\subsubsection{Planned-Outage Regime.}
\label{sssec:N-1_networks}
In the planned-outage regime, only the Greedy agent was applied, as the N-1 agent was designed to operate without outages present.
For data generation, the Greedy agent was applied to all scenarios of the planned-outage regime.

\begin{figure}[t]%
    \centering\includegraphics[width=0.8\textwidth]{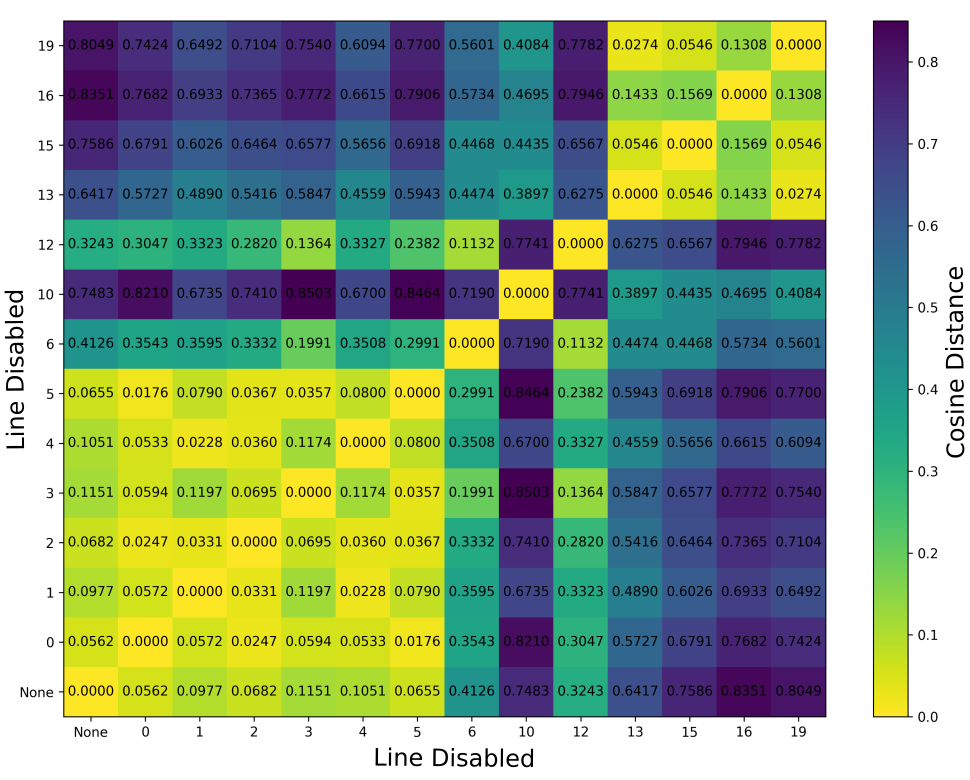}
    \caption{The cosine distance between the action distributions of greedy agents applied to the different (N-1) networks.
    The plot shows the presence of two clusters of (N-1) networks.
    The two yellow areas show the highly similar action distributions within these clusters.
    The blue areas show that the actions distributions are dissimilar between these two clusters.
    Because of the inability of the Greedy agent to operate them, certain N-1 networks are not included.}
\label{fig:cosine_distance_actions}%
\end{figure}

The performance of the Greedy agent depends greatly on which line is disabled, as described in Figure \ref{fig:rte_case14_realistic_annotated}.
Additionally, columns 3 and 4 of Table \ref{tab:agent_performance} show the performance of the Greedy agent with lines 2 or 6 disabled.
Two clusters emerge in the N-1 networks.
This is apparent in a comparison of the different action distributions in different N-1 networks, as shown in Figure \ref{fig:cosine_distance_actions}.
The first cluster consists of the N-1 networks with lines disabled in the transmission side of the grid and can be operated well by the Greedy agent (>96\% of days completed).
The second cluster concerns the lines that act as transformers and line 13, and is more difficult to operate by the Greedy agent (68-97\% of days completed).

The greedy agent could not be applied to one N-1 network as it is inherently invalid (line 18).
Additionally, the greedy agent could also not operate all N-1 networks well, also displayed in Figure \ref{fig:rte_case14_realistic_annotated}.
The agent failed almost all days with the disablement of a line in the distribution side of the grid (save for lines 10, 12, and 13).

\subsubsection{Unplanned-Outage Regime.}
As expected, the N-1 agent performs much better (92.39\% of days completed) than the greedy agent (81.79\% completed), as listed in Table \ref{tab:agent_performance}, column 5.
The results are only computed on the test scenarios as we do not generate data from this regime (due to the low performance of the greedy agent and the prohibitive application duration of the N-1 agent).

\section{Imitation Learning Agents}

\subsection{Dataset}
\label{ssec:dataset}
We create a dataset with the data from the N-1 agent (from the full-network regime) and data from the Greedy agent in the N-1 networks with lines 0, 2, 4, 5, 6, and 12 disabled.
We selected the lines that belong to the first cluster of N-1 networks as the N-1 agent performed well on these (see \S\ref{sssec:N-1_networks}).
We combine the data from the Greedy agent and the N-1 agent with the aim to train a ML agent that is capable of performing well in both the full and N-1 networks.
Explicit do-nothing data points and data points from failed days are excluded.
The data is split into train/validation/test sets based on the scenario they belong to (see \S\ref{ssec:scope_regime}).
The sets have 196,477/26,228/59,150 data points each. 

Each data point contains the features per object, the topology vector, and the expert action.
The features per object type are listed in Table \ref{tab:features}.
Object features are normalized, flattened into a vector, and concatenated with the topology vector. 
Features of the endpoints of disabled lines are zero-imputed. 
The corresponding value in the topology vector is -1.
Actions are transformed from a set-action format into a switch-action format.
As such, actions are represented by a vector with a length of the topology vectors, whose elements are either value zero or one, indicating whether the corresponding object is switched between busbars. 
This produces a multi-label binary classification task.

\begin{table*}[t]
\centering
\caption{Imitation learning features per object type.}
\label{tab:features}
\begin{tabular}{|l|l|l|l|} 
\cline{1-1}\cline{3-4}
\textbf{Generator/Load} & & \multicolumn{2}{l|}{\textbf{Line endpoint}} \\ 
\cline{1-1}\cline{3-4}
Active Production/Load &  & Active Power Flow & Current flow \\ 
\cline{1-1}\cline{3-4}
Reactive Production/Load &  & Reactive Power Flow & Loading ($\rho$) \\ 
\cline{1-1}\cline{3-4}
Voltage Magnitude &  & Voltage Magnitude & Thermal Limit \\ 
\cline{1-1}\cline{3-4}
\end{tabular}
\end{table*}

\subsection{Imitation Learning}

\subsubsection{Setup.}
We use a simple fully-connected neural network (FCNN) as the ML model.
The FCNN consists of an input layer, several hidden layers, and an output layer.
The output layer uses a sigmoid activation function to ensure the output is in the $(0, 1)$ range.
The hidden layers use the leaky ReLU activation function.
The weights are initialized according to a normal distribution, with the standard deviation as a tuned hyperparameter. We use the Adam optimizer to minimize a label-weighted binary cross-entropy loss, defined as:
\begin{align*}
L = &\textrm{mean}(-\textbf{w}(\textbf{y}\log(\textbf{p})+  (\textbf{1}-\textbf{y})\log(\textbf{1}-\textbf{p}))), 
\end{align*}
where $\textbf{p}, \textbf{y}$ and $\textbf{w}$ represent the prediction, target, and the label weight vectors respectively.
We introduced label weights because the sparsity of labels with value $1$ (since only a few objects switch per action) caused the predictions to collapse to only zeros.
The label weights assign lower weights to labels that do not belong to the target or predicted substation:
\begin{align}
w_i = \begin{cases}
1 & \text{object $i$ belongs to the target substation} \\
1 & \text{object $i$ belongs to the predicted substation} \\
\alpha & \text{otherwise}  
\end{cases} \quad \forall w_i \in \textbf{w} \nonumber
\end{align}
where $\alpha$ is the label weight hyperparameter. 
Note that there is potentially no target or no predicted substation, due to do-nothing actions.
The target substation, if any, depends on the data point's action.
There is no target substation if the action was an implicit do-nothing action.
The predicted action is considered a do-nothing action, without a predicted substation, if all predictions $p_i \in \textbf{p}$ do not exceed $0.5$. 
Otherwise, the predicted substation is the substation where the predictions $\textbf{p}_s$ at that substation $s \in S$ maximize
\begin{align*}
\Sigma_{p_i \in \textbf{p}_s} \textrm{max}(p_i-0.5,0).
\end{align*}

Each value in the output vector represents whether to switch the corresponding object.
However, not every output vector corresponds to a valid action.
We apply a postprocessing step where the model prediction $\textbf{p}$ is replaced by the valid switch-action nearest to $\textbf{p}$.
This postprocessing step is applied during validation, testing, and inference, but not during training.

\subsubsection{Training.}
Hyperparameter tuning was performed through multiple iterations of hyperparameter sweeps. 
Each sweep narrowed down the hyperparameter ranges and trained with more epochs and less strict early stopping.
The hyperparameter sweeps used random search with Hyperband early termination (distinct from early stopping), which terminates unpromising runs early.
Table \ref{tab:hyperparams} displays the hyperparameter ranges in the final sweep and the selected final values.
The weight decay was set to zero after observing that the best runs had values near the lower limit;
subsequent runs with a weight decay of zero performed even better.
We experimented with various values of the label weight $\alpha$ when we introduced label weighting, but did not include it in the hyperparameter sweeps.
The LReLU negative slope parameter was not tuned.

\begin{figure}[!t]
    \centering
    \includegraphics[width=0.6\textwidth]{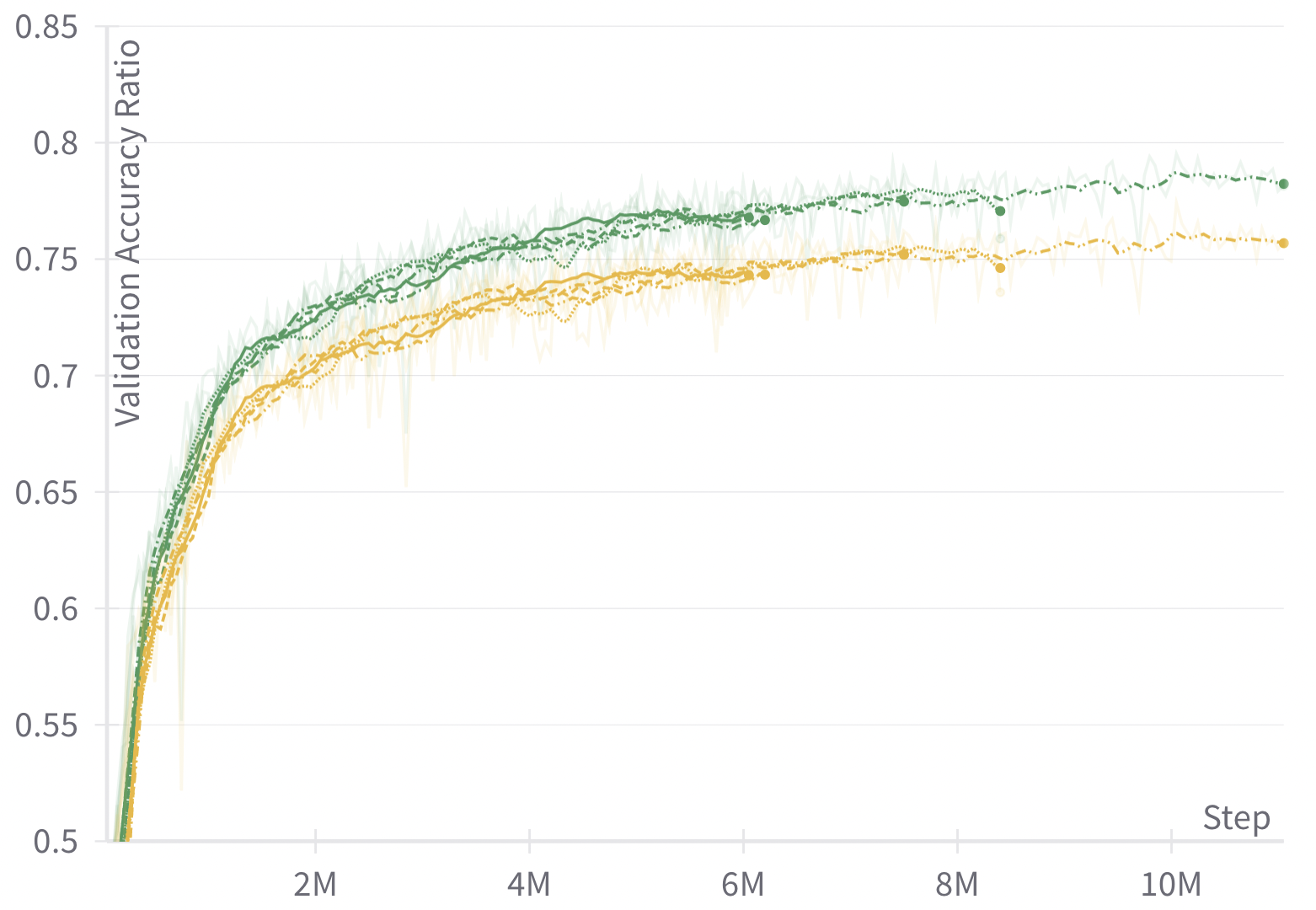}
    \caption{The training curves of the five final models.
    Green lines show the validation accuracy with the postprocessing step, yellow without.
    Lines are smoothed with a running average of 10.
    }
    \label{fig:training_curve}
\end{figure}

Five final models were trained, with different weight initializations. 
Each training run lasted one hundred epochs, unless it was stopped early.
Runs were stopped early if the maximum validation accuracy had not increased the last twenty evaluations. 
The validation accuracy was calculated every 250,000 iterations.
The training curves are displayed in Figure \ref{fig:training_curve}.

\begin{table}[t]
\centering
\caption{Hyperparameters, their selected values, and, if applicable, the ranges of their final hyperparameter sweeps.}
\label{tab:hyperparams}
\begin{tabular}{|l|l|l|l|l|l|l|}
\cline{1-3} \cline{5-7}
\textbf{Hyperparameter} & \textbf{Value} & \textbf{Range} & & \textbf{Hyperparameter} & \textbf{Value} & \textbf{Range}  \\ 
\cline{1-3} \cline{5-7} \rule{0pt}{2.2ex} \#Hidden layers  & 4 & (1, 5) & & Weight decay  & 0 & $(e^{-9}, e^{-2})$ \\ 
\cline{1-3} \cline{5-7}\rule{0pt}{2.2ex} \#Hidden nodes & 230 & (32, 256) & & Weight init. $\sigma$ & 5 & $(e^{-0.5}, e^{1.5})$ \\ 
\cline{1-3} \cline{5-7} Batch size & 64 & (32, 128) & & Label weight $\alpha$ & 0.1 & -\\ 
\cline{1-3} \cline{5-7} \rule{0pt}{2.2ex}Learning rate & 7E-4 &($e^{-9}, e^{-5})$ & & LReLU neg. slope & 0.1 & - \\ 
\cline{1-3} \cline{5-7} 
\end{tabular}
\end{table}

\subsection{Imitation Learning Agents}

The obtained FCNN models can be applied to environment directly, but we are also interested in the performance of agents that combine ML and simulation capabilities. 
We evaluate four agents that use the ML model.

The \textit{Naive agent} executes the predicted action directly.
We observe that the Naive agent occasionally failed days by predicting a singular bad action.
The \textit{Verify agent} addresses this by verifying predicted action with simulation.
If the simulated maximum loading of the action exceeds the current max loading and the risk threshold parameter $\riskthreshold = 1$, the action is replaced by a do-nothing action.
We also consider hybrid agents.
The \textit{Verify+Greedy agent} functions as the verify agent as long as the maximum loading is below the risk threshold and as the Greedy agent otherwise.
The \textit{Verify+N-1 agent} functions identically, except with N-1 agent instead of the Greedy agent.
Each of these agents uses the activity threshold $\dnthreshold = 0.97$ for active behavior.

\subsection{Results}

\subsubsection{IL Model.}
For the five models, the accuracies on the training, validation, and test sets were $79.1\% \pm1.5$, $78.6\%\pm0.6$, and $76.4\%\pm 0.6$, respectively. 
The action postprocessing step had a sizeable effect on accuracy: the validation accuracy without this step decreased from $78.6\%\pm0.6$ to $76.2\% \pm0.7$ (see Figure \ref{fig:training_curve}).

The accuracies are limited to approximately 80\%.
The class imbalance seems to contribute to the limited accuracy, as indicated by the negative relation between class rank and accuracy shown in Figure \ref{fig:accuracy_per_label}.
Figure \ref{fig:label_predictions} indicates that the model predicts infrequent classes disprortionally infrequently.
However, the class imbalance alone seems insufficient to explain the limited accuracy, as shown by the high variance in accuracy for actions with similar frequencies, in Figure \ref{fig:accuracy_per_label}.
Additionally, the accuracy is limited even on frequent classes: the accuracy on the fourth and fifth most frequent classes is $82.3\%$ and $61.1\%$ respectively.

\begin{figure}[!t]%
    \centering
    \subfloat{{\includegraphics[width=6cm]{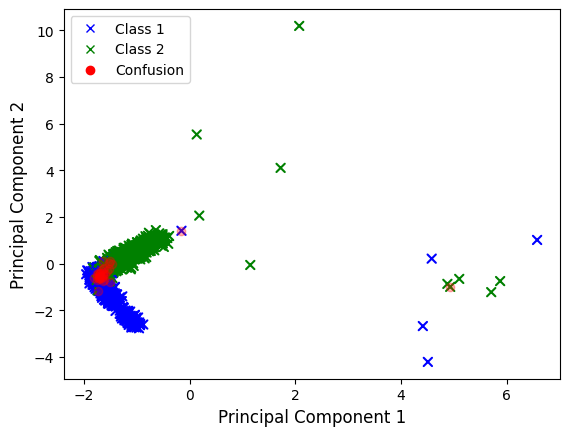}}}
    \subfloat{{\includegraphics[width=6cm]{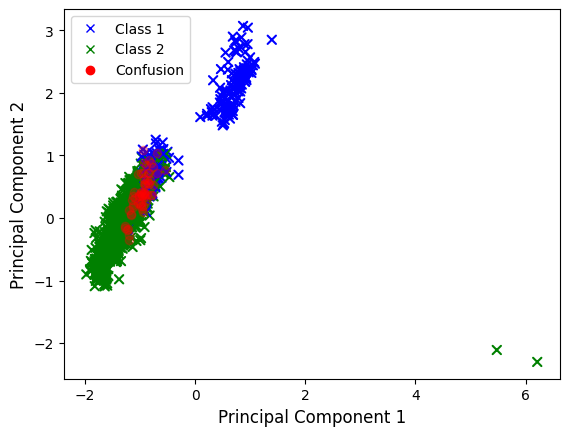} }}%
    \caption{The data points of the most confused classes projected on the first two principal components, for the N-1 networks with line 0 (left) and line 2 (right) disabled.
    The confused data points are overlaid in red.}
    \label{fig:confusions}
\end{figure}

Class overlap also appears to contribute to the limited accuracy.
Per (N-1) network, we found the most frequently confused pairs of classes and performed a principal components analysis on the features.
We also transformed the actions back from a switch-action format to a set-action format. 
In Figure \ref{fig:confusions}, we plot the data points of the confused classes projected on the first two principal components, highlighting which data points are confused.
As is visible, the confusions occur exactly where the classes overlap (red points occur at the intersection of the blue and green points).
The pattern is consistent across (N-1) networks and pairs of confused classes.
An inspection of the nearest neighbors supports the notion that the class overlap causes the low accuracy.
For a subset of 2500 validation data points, the accuracy over the data points whose nearest neighbors is in the same class is 92.97\%, but only 44.44\% for those whose nearest neighbor is in another class.

\subsubsection{Analysis of IL Agents.}
Table \ref{tab:agent_performance} shows the resulting performances of the agents on the different environmental regimes (explained in \S\ref{ssec:scope_regime}). 
A partial ordering holds true in all settings: Do-nothing < Naive < Verify < Verify + Greedy < Verify + N-1.
Among the ML agents, performance increases with more simulation in every setting (rows 5 to 8).
In every setting, the Verify agent improves upon the Naive agent by one to three percent (rows 5 and 6), whereas the improvements by the hybrid agents are more variable (rows 6, 7, and 8).
In the full-topology regime, the hybrid agents approach the performance of the N-1 agent (column 2).
Moreover, in the planned-outage and unplanned-outage regimes, hybrid agents fully achieve the performance of both rule-based agents (columns 3 to 6).

\begin{figure}[t]%
    \centering
    \subfloat{{\includegraphics[width=6cm]{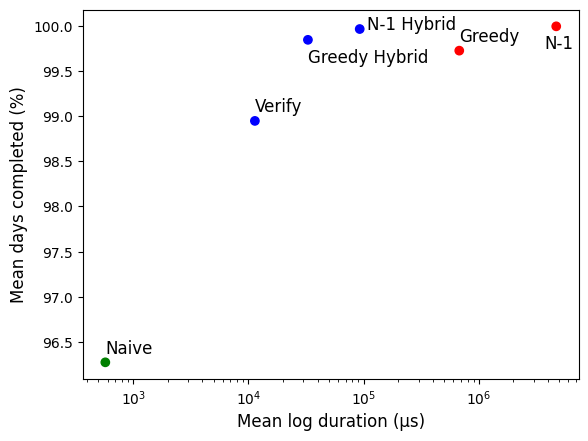}}}%
    \subfloat{{\includegraphics[width=5.8cm]{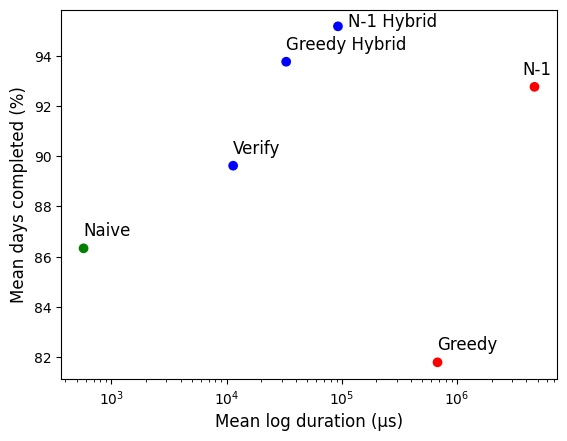}}}%
    \caption{The performance and inference duration for the agents.
    The combinations of speed and performance of the hybrid agents are favorable in comparison to the rule-based agents.
    Left: full-topology regime. Right: unplanned-outage regime.}
    \label{fig:tradeoff}
\end{figure}

Finally, we investigated the inference speed\footnote{The inference speed was measured on an apple M1 CPU with minimal background processes, on the first fifty validation scenarios.} of the different agents.
Figure \ref{fig:tradeoff} plots the performance and inference duration of each agent (values listed in Table \ref{tab:agent_performance}, column 6).
The mean inference speed of the IL agents is orders of magnitude smaller than those of the rule-based agents, even for the hybrid agents.
Consequently, the hybrid agents can achieve performance similar to the rule-based agents at a much higher inference speed.

\section{Summary \& Discussion}
Topology control is a underutilized opportunity for managing congestion in power grids.
Imitation learning is a promising technique for topology control, but its application has not been studied in detail.
In this research, we train FCNN models on the state-action pairs of two rule-based expert agents, and consequently apply these model to the power grid as  agents.

We find that the rule-based agents, particularly the N-1 agent, can effectively control the grid in an intra-day setting without outages.
The performance of both agents decreases in the unplanned-outage regime, but far more so for the greedy agent.
We also find that the N-1 agent performs longer action sequences.

We successfully trained FCNN models with imitation learning.
We introduced an action post-processing step that contributed to a small increase in accuracy.
However, the accuracies remain limited due to an imbalanced class distribution and class overlap.
We hypothesize that the class overlap results from usage of the Grid2Op \texttt{simulate} function by the agents.
The \texttt{simulate} function simulates the next timestep with generator/load forecasts predetermined by the scenario.
It's possible that (near-)identical network states have diverging forecast that lead to the selection of different actions.

The purely ML agents are adept as operating the grids in the different regimes, although the hybrid agents do outperform them.
This is in line with the findings of the L2RPN 2020 submission analysis \cite{marot2021learning}. 
The most unexpected result is that the hybrid agents outperform the rule-based agents in the planned-outage and unplanned-outage regimes, which cannot be explained through an IL lens.

We hypothesize that the hybrid models can find more robust topologies than the rule-based models due to model bias against rare actions.
As discussed in section \S\ref{ssec:full_network_regime}, the action distribution is highly imbalanced.
The few dominant actions are selected often by the rule-based agents, indicating that these action result in optimal topologies across many grid states. 
Their optimality in many grid states indicates their robustness.
In contrast, rare actions are optimal in only few grid states, potentially making them over-specialized. 
Figure \ref{fig:label_predictions} shows that the model predicts rare classes disproportionally infrequent.
This bias towards more frequent actions might consequently bias the model towards more robust, and consequently, more beneficial, topologies.

The ML and hybrid agents have good combinations of performance and inference speed.
All ML and hybrid agents represent a different tradeoff between performance and speed.
In practice, the optimal agent choice depends on the available resources and performance requirements.

\section{Future Work}

Various directions stand out as promising for future research. 
There is as yet little knowledge about the best practices for creating a dataset or training a model with IL for topology control.
Future research might shed light on such best practices.
The class overlap present in our dataset might be avoided by including the forecasted production and consumption in the features.
The developed IL methods must be compared against RL methods. 

Simple behavioral cloning, as researched here, can result in \textit{distribution shift}: a compounding divergence between the behavior of the expert and student agent \cite{pmlr-v15-ross11a}.
This might have been exasperated in the unplanned-outage setting, where agents were evaluated on a regime not included in the dataset.
It seems promising to apply more advanced imitation learning frameworks.
DAgger, for instance, mitigates the aforementioned distribution shift problem \cite{pmlr-v15-ross11a}. 
An in-depth analysis of the behavior of rule-based, purely ML, and hybrid agents would be interesting, as it might clarify what behavior leads to good performance.

There remains a large divergence between environments studied in L2RPN competitions and real power grids.
Real grids are larger, more complex (e.g. including other grid objects and varying number of busbars between substations), and permit other types of actions.
They are also frequently subject to changes, which means that models should generalize to unseen networks.
Research is required to integrate these complications with existing methods.
Graph neural networks seem promising w.r.t. generalizability.

\begin{credits}
\subsubsection{\ackname} AI4REALNET has received funding from European Union’s Horizon Europe Research and Innovation programme under the Grant Agreement No 101119527. Views and opinions expressed are however those of the authors only and do not necessarily reflect those of the European Union. Neither the European Union nor the granting authority can be held responsible for them.
\end{credits}

\bibliographystyle{splncs04}


\end{document}